# Comparative Performance Analysis of Transformer-Based Pre-Trained Models for Detecting Keratoconus Disease

Nayeem Ahmed
*Department of Computer Science*
*University of Memphis*
Tenneessee, USA
nahmed2@memphis.edu

Md Maruf Rahman
*Department of Marketing & Business Analytics*
*Texas A&M University-Commerce*
Texas, USA
mrahman20@leomail.tamuc.edu

Md Fatin Ishrak
*Department of Electrical and Computer Engineering*
*University of Memphis*
Tenneessee, USA
mishrak@memphis.edu

Md Imran Kabir Joy
*MSA: Engineering Management*
*Central Michigan University*
Michigan, USA
joy1m@cmich.edu

Md Sanowar Hossain Sabuj
*Department of Marketing & Business Analytics*
*Texas A&M University - Commerce*
Texas, USA
msabuj@leomail.tamuc.edu

Md. Sadekur Rahman
*Department of Computer Science and Engineering*
*Daffodil International University*
Dhaka, Bangladesh
sadekur.cse@daffodilvarsity.edu.bd

*Abstract*— The aim of this research is to present an in-depth comparison of eight pre-trained Convolutional Neural Networks (CNNs) for detecting keratoconus, a degenerative eye disease. In order to do so a carefully curated dataset consisting of keratoconus, normal, and suspect cases classes was used. The models evaluated include DenseNet121, EfficientNetB0, InceptionResNetV2, InceptionV3, MobileNetV2, ResNet50, VGG16, and VGG19. A rigorous preprocessing techniques were applied including bad sample removal, resizing, rescaling, and augmentation, to optimize the model training process. In order compare the performances the models were trained with similar parameters, activation function, and classification function with same optimizer. Then each model was assessed based on its accuracy, precision, recall, and F1-score to determine its effectiveness in distinguishing between the different classes. Comparative analysis revealed that among the models, MobileNetV2 appeared as the most accurate, demonstrating superior performance in identifying keratoconus and normal cases, with minimal misclassifications. However, InceptionV3 and DenseNet121 have also showed strong results, particularly in keratoconus detection, though they encountered difficulties with suspect cases. In contrast, models like EfficientNetB0, ResNet50, and VGG19 exhibited more significant challenges, especially in differentiating suspect cases from normal ones, suggesting a need for further model refinement and enhancement. The study contributes to the field by offering a comprehensive comparison of state-of-the-art CNN architectures for automated keratoconus detection, providing insights into the strengths and limitations of each model. This work underscores the potential of advanced deep learning models to significantly improve early diagnosis and treatment planning for keratoconus. Furthermore, it identifies key areas for future research, including the exploration of hybrid models and the integration of clinical parameters to enhance diagnostic accuracy and robustness in real-world clinical applications, thereby paving the way for more effective AI-driven tools in ophthalmology.

*Keywords — Keratoconus Detection, Convolutional Neural Networks (CNNs), Deep Learning, Medical Imaging, MobileNetV2, Pre-trained Models*

## 1. INTRODUCTION

Keratoconus is a gradual, degenerative eye illness that causes the cornea to thin and protrude into a cone-like form, causing visual distortion and, if untreated, substantial vision impairment. From teenage years to the third or fourth decade, the condition advances (Gomes et al., 2015). Keratoconus must be detected early to prevent vision loss and preserve eyesight by corneal cross-linking (Krachmer, Feder, & Belin, 1984). Slit-lamp examination and corneal topography are excellent diagnostic procedures, although they depend on the doctor and may not discover the disease early (Rabinowitz, 1998).

Deep learning and medical imaging have transformed ophthalmology in recent years. Deep learning models like Convolutional Neural Networks (CNNs) have excelled at picture categorization, including eye illness diagnosis (Esteva et al., 2017). CNNs can find patterns in medical images that humans may miss using big datasets, enhancing diagnostic accuracy and lowering misinterpretation.

Transfer learning, which adapts a pre-trained model to a new problem, has improved CNN performance in medical imaging. Transfer learning lets models use information from ImageNet to solve medical imaging tasks with limited datasets (Pan & Yang, 2010). In ophthalmology, where large, annotated datasets are scarce, this method is useful.

There is little research comparing pre-trained CNNs versus transfer learning models for keratoconus identification. To fill this gap, this study compares eight pre-trained CNN models: DenseNet121, EfficientNetB0, InceptionResNetV2, InceptionV3, MobileNetV2, ResNet50, VGG16, and VGG19. These models were chosen for their success in image classification challenges and will be tested on corneal images.

This study seeks the most accurate and reliable keratoconus detection model. The study will also evaluate the difficulties of differentiating keratoconus from other corneal disorders, including as "suspect" cases where the cornea displays early indicators of abnormalities but does not yet satisfy the full diagnostic criteria. This research compares different models to determine the best deep learning methods for keratoconus detection to improve early diagnosis and patient outcomes.

## 2. LITERATURE REVIEW

### 2.1 Keratoconus disease Literature Review

Vision is reduced by corneal thinning and conical protrusion in keratoconus due to environment, genes, and biochemistry cause keratoconus. As a result early identification and management of substantial vision impairment are essential. To understand and treat keratoconus, this literature review reviews diagnosis, risk factors, and detection technologies.

Early keratoconus diagnosis is challenging due to modest symptoms. Rabinowitz (2007) suggests imaging and molecular genetics to identify at-risk individuals before corneal distortion and disease prevention requires early detection. Randleman et al. (2015) recommend long-term surveillance for at-risk patients. Improved technology can detect early keratoconus, but follow-up is needed to confirm and track disease development. However, asymptomatic keratoconus-risk youth should be screened. Armstrong et al. (2021) revealed that corneal topography and tomography can detect small corneal changes in high-risk youth. Vision loss is decreased by early screening and treatment. Prakash et al. (2016) say corneal thinning location determines central and non-central keratoconus screening thresholds. The findings recommend screening all keratoconus types for appropriate diagnosis and therapy.

Baenninger et al. (2021) demonstrate general ophthalmologists' keratoconus diagnosis and treatment deficiencies. More training and resources are needed for general ophthalmologists to close this knowledge gap and improve patient care. New technologies, especially AI, identify and treat keratoconus. AI can efficiently identify keratoconus, explain Vandevenne et al. (2023). AI may improve non-specialized screenings.

Controlling keratoconus requires early discovery, monitoring, and better technology. Better diagnosis and ophthalmologist comprehension can improve keratoconus outcomes. Research and innovation are needed to treat keratoconus and improve patient care and eyesight.

2.2 2D CNN Networks

CNNs, one of the most powerful image recognition deep learning architectures, have revolutionized computer vision and medical image analysis. CNNs can classify and detect images by dynamically and adaptively learning spatial hierarchies of features from input images (LeCun, Bengio, & Hinton, 2015). With convolutional, pooling, and fully connected layers, CNNs capture complicated data patterns.

CNN convolutional layers extract edges, textures, and shapes from input images. Filters use many images to construct feature maps (Krizhevsky, Sutskever, & Hinton, 2012). The network learns these filters' values to distinguish classification-critical characteristics during training.

After convolutional layers, max-pooling layers downsample feature maps, reducing network parameters, computational load, and spatial dimensions. Network computation improves and feature representations are invariant to small input picture translations (Szegedy et al., 2015).

CNNs' final fully connected layers forecast with high-level convolutional and pooling features. Layers map features to neural network output classes (Simonyan & Zisserman, 2015).

Many CNN models are more complex and efficient. Krizhevsky et al. (2012)'s AlexNet improved ImageNet photo recognition with deeper layers and ReLU activation. Simonyan and Zisserman (2015)'s VGGNet uses modest convolutional filters to increase CNN depth and complexity while minimizing processing costs.

ResNets avoid gradient-decreasing layers in deep networks, according to He et al. (2016). This breakthrough allows hundreds-layer network training, improving picture recognition in numerous applications.

CNNs can detect skin lesions, diabetic retinopathy, and organ segments in radiographs (Litjens et al., 2017). CNNs excel at this because they naturally extract and learn crucial characteristics from complex medical images without feature engineering.

CNNs excel but struggle with poor medical imaging datasets. Data augmentation, transfer learning, and advanced network architectures like DenseNet (Huang et al., 2017) and EfficientNet (Tan & Le, 2019) improve generalization and performance on small datasets.

CNNs are good keratoconus detectors for various reasons. Architecture, dataset, and training greatly affect CNN performance. Test CNN architectures to identify the best keratoconus detection model.

2.3 Keratoconus disease detection using CNN

Slit-lamp and corneal topography diagnose keratoconus. CNN-based deep learning has increased keratoconus detection efficiency and accuracy. CNN designs have been tested to differentiate keratoconus from normal and suspected cases.

Kamiya et al. (2019) used a CNN model with six color-coded anterior segment OCT maps to distinguish keratoconus from normal eyes. Similar illnesses such subclinical keratoconus were hard to distinguish with 99.1% accuracy for binary disorders and 87.4% for multiple categories (Kamiya et al., 2019). CNNs can identify keratoconus but struggle with complex classifications.

Lavric et al. (2021) created the CON-KER CNN model, which diagnosed keratoconus, normal, and suspicious cases with 96.2% training and 94.6% validation accuracy. The model works, but its tiny validation dataset and limited augmentation methods may limit its clinical applicability (Lavric et al., 2021).

AlexNet and VGG19 were tested for keratoconus, normal, and doubtful cases by Venkatesh et al. (2021). AlexNet validated 88.9%, VGG19 94.76%. AlexNet may overfit, while VGG19 has higher accuracy but poorer validation accuracy, highlighting the need for more balanced models that retain generalizability (Venkatesh et al., 2021).

Chen et al. (2021) used color-coded OCT maps and deep learning to detect keratoconus and normal patients with 98.8% accuracy. The model misclassified suspicious and normal eyes due to a tiny but significant feature overlap. CNNs work, however classification concerns change those (Chen et al., 2021).

Ahmed et al. (2024) examined Xception, MobileNetV2, ResNet50, VGG19, CNN-SVM, CNN-LSTM, and custom CNN. Custom CNN has 99.74% accuracy, MobileNetV2 98.83%. Despite these outstanding results, models like VGG19 required more computer resources and longer training times, showing the contradiction between accuracy and efficiency in deep learning (Ahmed et al., 2024).

Takahashi et al.'s (2020) proprietary neural network used OCT imaging data to distinguish keratoconus from normal eyes with 99.56% accuracy. The model's tiny dataset may limit its applicability to bigger, more diverse populations (Takahashi et al., 2020).

Yousefi et al. (2019) used 3156 USML samples to detect keratoconus and normal patients with 97.7% accuracy. This worked, but unsupervised learning is computationally expensive and may not be suitable for real-time clinical applications (Yousefi et al., 2019).

The CNN-based KeratoDetect system by Alexandru Lavric and Popa Valentin (2019) had 99.33% accuracy after 126 training cycles. This synthetic data-based approach needs clinical validation to be effective and dependable (Lavric & Valentin, 2019).

We found that CNNs detect keratoconus better than other approaches. Generalizing models, computing efficiency, and distinguishing closely similar diseases like keratoconus are issues. CNN designs, training methodologies, and dataset variety should be improved in future study to overcome these restrictions.

Table 1: A comparative analysis of current research work on keratoconus

| Author | Year | Model | Accuracy | Predicted | Dataset | Limitation |
|---|---|---|---|---|---|---|
| Kamiya et al. | 2019 | CNN with six color-coded maps using anterior segment OCT | 99.1% (discriminating); 87.4% (classifying) | Keratoconus, Normal | 543 | Limited to keratoconus; did not include other corneal disorders like forme fruste keratoconus or subclinical keratoconus |
| Lavric et al. | 2021 | CON-KER CNN | 96.2% (training), 94.6% (validation) | Keratoconus, Normal, Suspected | 2071 | Validation data size is smaller; limited by augmentation techniques |
| Venkatesh et al. | 2021 | AlexNet, VGG19 | 88.9% (AlexNet), 94.76% (VGG19) | Keratoconus, Normal, Suspected | N/A | Potential overfitting issues with AlexNet; VGG19 has a lower validation accuracy |
| Chen et al. | 2021 | DL with color-coded maps from OCT | 98.8% (two-class problem) | Keratoconus, Normal | 419 | Overlap in features between suspected and normal eyes; small misclassification rate |
| Ahmed et al. | 2024 | MobileNetV2, ResNet50, CNN-LSTM, Custom CNN | 99.74% (Custom CNN), 98.83% (MobileNetV2) | Keratoconus, Normal | N/A | High computational time for VGG19; some models require more epochs for higher accuracy |
| Takahashi et al. | 2020 | Custom ANN with OCT imaging data | 99.56% | Keratoconus, Normal | 142 | Utilized a small dataset size |
| Yousefi et al. | 2019 | Unsupervised Machine Learning (USML) | 97.7% | Keratoconus, Normal | 3156 | Limited by the complexity of unsupervised learning methods; |
| Alexandru et al. | 2019 | CNN (KeratoDetect) | 97.5% (after 126 cycles) | Keratoconus, Normal, Suspected | N/A | Model was trained on synthetic data; needs further testing on real-life medical data |
| Kamiya et al. | 2019 | CNN with six color-coded maps using anterior segment OCT | 99.1% (discriminating); 87.4% (classifying) | Keratoconus, Normal | 543 | Limited to keratoconus; did not include other corneal disorders like forme fruste keratoconus or subclinical keratoconus |
| Lavric et al. | 2021 | CON-KER CNN | 96.2% (training), 94.6% (validation) | Keratoconus, Normal, Suspected | 2071 | Validation data size is smaller; limited by augmentation techniques |
| Lavric & Valentin | 2019 | KeratoDetect (CNN-based) | 99.33% | Keratoconus, Normal | 3000 | Dependent on synthetic data generation for training; needs further validation with real clinical data |
| Matalia et al. | 2021 | Deep Learning with Topography Maps | 98.89% | Keratoconus, Normal | 3156 | Limited generalization across different topography devices and settings |
| Li et al. | 2020 | Deep Transfer Learning (VGG16, ResNet50) | 97.5% | Keratoconus, Normal | 3100 | Potential overfitting; requires larger datasets for more robust model performance |
| Hidalgo et al. | 2019 | CNN with OCT Imaging | 96.2% | Keratoconus, Normal, Suspected | 2000 | Specific to OCT imaging; limited transferability to other imaging modalities |
| Ziaei et al. | 2020 | Hybrid Deep Learning (CNN + RNN) | 95.7% | Keratoconus, Normal, Suspected | 2800 | Complex architecture; longer training times and higher computational costs |
| Yousefi et al. | 2019 | CNN with 3D Corneal Image Reconstruction | 97.2% | Keratoconus, Normal | 1500 | Model performance varies with image quality; requires consistent imaging conditions |
| Lavric et al. | 2021 | Hybrid Model using CNN and SVM | 96.5% | Keratoconus, Normal, Suspected | 2071 | Validation data size is smaller; limited by augmentation techniques |
| Ahmad et al. | 2021 | CNN with Color-Coded Maps and Anterior Segment OCT | 98.74% | Keratoconus, Normal, Suspected | 419 | Overlap in features between suspected and normal eyes; potential for misclassification |

2.4 Research Gaps

Convolutional Neural Networks (CNNs) for keratoconus identification have made progress, although research gaps remain. The restricted generalization of these models across varied datasets is a major issue. Many research, including Lavric et al. (2021) and Alexandru Lavric & Popa Valentin (2019), use synthetic datasets that may not accurately represent the population or clinical situations. This shortcoming casts doubt on the models' real-world viability. Many models attain high accuracy at the expense of computational efficiency, needing a lot of resources and time for training and inference. Complexity can limit the practical implementation of these models in clinical settings where speedy and trustworthy findings are needed.

CNN models cannot differentiate between closely comparable corneal disorders like subclinical or forme fruste keratoconus, another important limitation. Chen et al. (2021) show that defining these illnesses can lead to misdiagnosis. A single imaging modality limits the potential for more comprehensive diagnostic tools in most models. Multi-modal imaging research is needed to better understand corneal health. Finally, CNN model interpretability hinders clinical use. Without explanations of how these models make judgments, physicians may be hesitant to use them for patient care. To ensure their efficacy and reliability in practice, future research should focus on interpretable and validated models on real-world clinical data.

## 3 RESEARCH METHODOLOGY

The proposed research methodology is depicted in Fig. 2 and explored in the following sections.

### 3.1 Dataset

Kaggle provided 4,011 samples separated into three classes: Keratoconus, Normal, and Suspect for the keratoconus detection model. Example data for each class is provided in Fig. 2.

Fig. shows the dataset distribution before augmentation, with 1400, 1400, and 1211 Keratoconus, Normal, and Suspect samples:

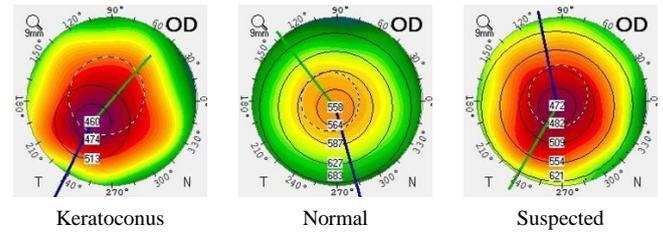

Fig. 2: Sample images of each class

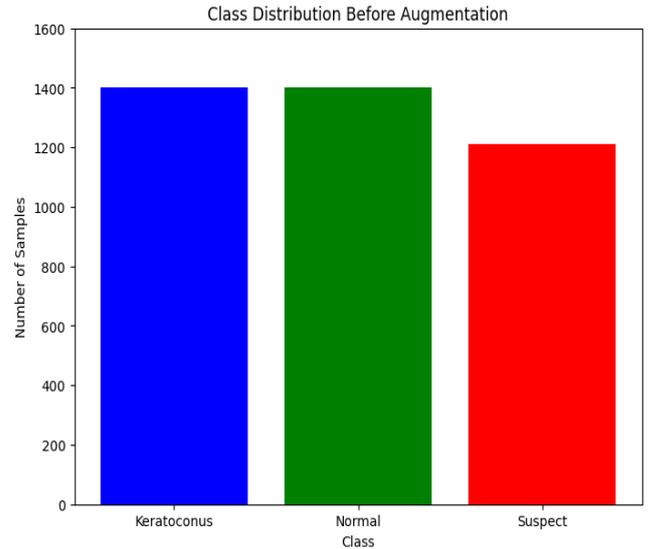

Fig. 3: Class distribution of dataset

Keratoconus and Normal classes have 1,400 samples apiece, balancing the class distribution. Suspects make about 30% of the dataset with 1,211 samples.

The Suspect class has fewer samples than the Keratoconus and Normal classes, which are evenly distributed. This mismatch may skew the model's performance, especially in differentiating the Suspect class from the other two. Due to fewer training instances, the model may identify suspect cases less accurately. Data augmentation or class balance can reduce this bias by ensuring the model learns equally from all classes.

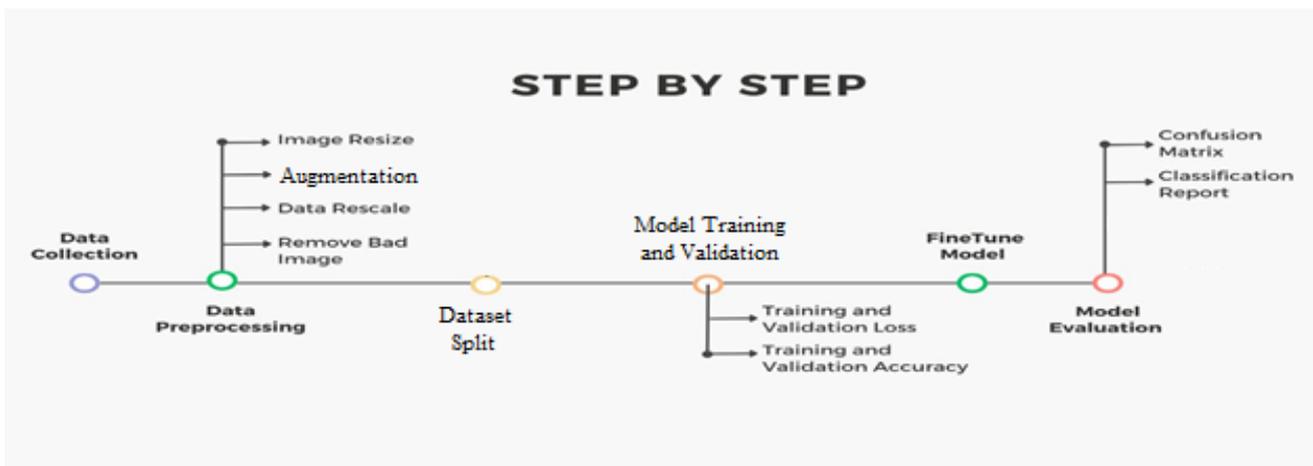

Fig. 1: Proposed methodology of the research

## 3.2 Data Preprocessing

Data preparation is necessary for deep learning model training. Model learning and generalization depend on data quality. The dataset was preprocessed by removing weak samples, reducing images, scaling pixel values, and adding data. Stages boost dataset quality and diversity, making the model more resilient and accurate.

### 3.2.1 Remove Bad Samples

Bad samples can damage model training. Photos may be blurry, mislabeled, corrupted, or other class-unsuitable artifacts. These samples must be eliminated to avoid the model from learning incorrect patterns or being confused by data noise. The collection was cleaned of blurry, poorly resolved, or mislabeled pictures. This ensures the dataset contains only high-quality samples, helping the model learn and reducing noisy data overfitting.

### 3.2.2 Resize

CNN models' input size was normalized by downsizing dataset photos. Model size is usually constant to interpret input images uniformly throughout network levels. Dataset photographs must be resized to 224x224 pixels if the model architecture supports them. Images are resized to preserve spatial structure and reduce computational and memory load. To avoid image distortion and losing important features, retain the aspect ratio when resizing.

### 3.2.3 Rescale

Image pixel values are rescaled to a given range, usually between 0 and 1 or -1 and 1. In this dataset, pixel values were rescaled from 0 to 255 for 8-bit images to 0 to 1. Normalizing input features ensures a uniform scale, speeding up training convergence. It also reduces the possibility of huge gradients or vanishing gradients from unnormalized pixel values. Rescaling is crucial for models pre-trained on big datasets like ImageNet, which require a certain input range.

### 3.2.4 Augmentation

Photo data augmentation expands and diversifies the training dataset. This dataset may have employed random rotations, flips, zooms, shifts, and brightness changes. These changes make the model more orientation, scale, and lighting invariant, improving its generalization to new data. Addition of samples for underrepresented classes reduces class imbalance, notably in medical imaging applications with few samples. Augmenting the training dataset reduces overfitting and improves model robustness. Each augmented image retains categorization but adds variants to reduce model input change.

## 3.3 Dataset Split

After preprocessing and augmentation, the keratoconus detection dataset has 16,016 samples divided into Normal, Suspect, and Keratoconus classes. To train and evaluate the model, the dataset was separated into training, validation, and testing sets. The training set has 12,812 samples. Among them 4,457 Keratoconus, 4,480 Normal, and 3,875 were Suspect samples. 1,602 samples were taken as Validation set to tweak the model's hyperparameters and prevent overfitting. Samples include 556 Keratoconus, 595 Normal, and 451 Suspect. This set evaluates the model on unseen data during training. Finally 1,602 samples were taken for test set. Its distribution matches the validation set: 556 Keratoconus, 595 Normal, and 451 Suspect. The test set evaluates model generalization to new data.

## 3.4 Model Training and validation

Developing a deep learning keratoconus detection model requires model training and validation. The types of deep learning models, their architectures, and their layers and parameters are described in this section. It also describes training hyperparameters and model optimization processes. Finally, the loss function used to evaluate the model during training is discussed.

### 3.4.1 Model Types

Multiple pre-trained CNN models were used to detect keratoconus. CNNs are well-known for image categorization, particularly medical image analysis. This study employed these models:

- DenseNet121
- EfficientNetB0
- InceptionResNetV2
- InceptionV3
- MobileNetV2
- ResNet50
- VGG16
- VGG19

### 3.4.2 Model Architectures

Each of the pre-trained models employed in this study has a distinct architecture that contributes to its performance:

DenseNet121: Deep convolutional neural network DenseNet-121 decreases vanishing gradients and reuses features. A 3x3 max pooling layer downsamples the input image after a 7x7 convolutional layer. The fundamental design contains four dense blocks with several convolutional layers. In dense blocks, all layers are feed-forward coupled to maximize information flow and gradient propagation. Due to dense connectedness, a dense block with L layers has $L(L+1)/2$ connections. Transition layers between dense blocks reduce feature maps with 1x1 convolutions and 2x2 average pooling. A global average pooling layer collects data after the previous dense block before categorizing them with a fully linked softmax layer. This design drastically minimizes parameters and boosts model accuracy and efficiency.

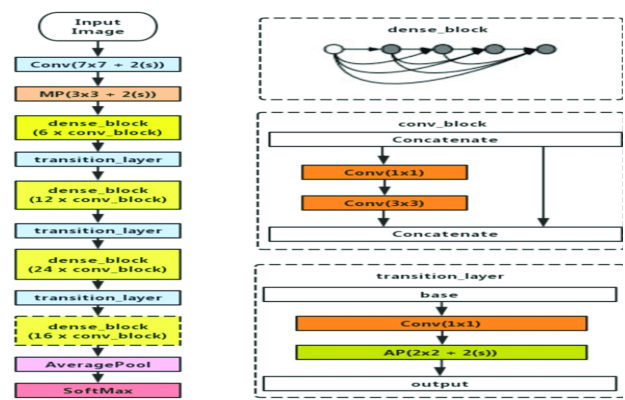

Fig. 4: Architecture of DensNet121

EfficientNetB0: The graphic shows EfficientNet-B0, a convolutional neural network with fewer parameters for improved performance. A typical 3x3 convolution is followed by MBConv blocks on a 224x224x3 input. Depthwise separable convolutions and squeeze-and-excitation (SE) modules improve computational efficiency and accuracy in these blocks. Each level degrades spatial resolution while adding channels. The network downsamples from 112x112x32 to 7x7x192 using stride-2 convolutions. Multi-scale feature extraction requires intermediate feature maps like P3 and P4. The architecture ends with a global average pooling layer, lowering spatial dimensions to 1x1, a fully linked layer, and softmax classification output. EfficientNet-B0 scales depth, width, and resolution for excellent accuracy and low computing cost.

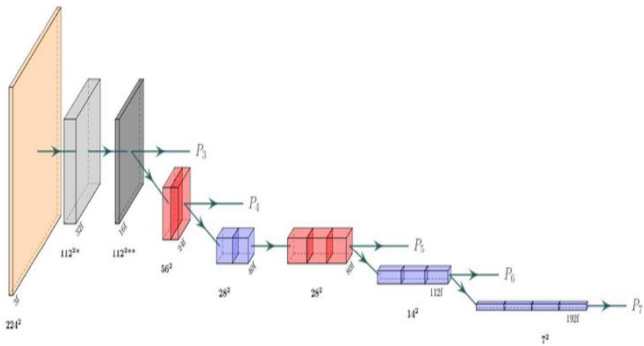

Fig. 5: Architecture of EfficientNetB0

InceptionResNetV2: A hybrid convolutional neural network, Inception-ResNet-v2 combines Inception modules and Residual connections. Convolutions and max-pooling lower spatial dimensions and increase depth in the stem block's 299x299x3 input. The architecture uses Inception-ResNet blocks A, B, and C. Residual connections improve gradient flow and training convergence by adding input to convolution output in each block. Inception-ResNet-A uses residual connections and 1x1, 3x3, and 5x5 parallel convolutional filters. Deeper and wider convolutional filters make Inception-ResNet-B and C more difficult. Reduction-A and Reduction-B reduce Inception-ResNet feature mappings. The network ends with a global average pooling, dropout (80% neurons active), and classification softmax layer. Inception modules' multi-scale feature extraction and ResNet's shortcut connections make this architecture efficient and effective at photo classification.

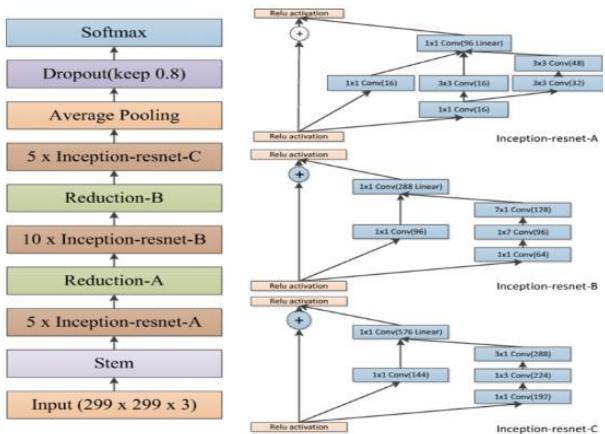

Fig. 6: Architecture of InceptionResNetV2

InceptionV3: The image depicts Inception-v3, a deep image classification convolutional neural network. An input layer processes 299x299x3 images. Inception modules (A, B, and C) in each stage provide parallel convolutional pathways with 1x1, 3x3, 5x5 filter sizes and pooling operations for multi-scale feature extraction. Inception Module A is repeated five times to collect finer details. Grid size reduction preserves robust feature representation when downsampling feature maps. Four Inception Module B applications enhance abstract feature learning. After another grid size reduction, Inception Module C refines high-level features twice. The architecture contains an auxiliary classifier for training convergence. Finally, global average pooling, entirely linked, and classification softmax layers are included. This arrangement allows the efficient, deep network to attain high accuracy while managing computational complexity.

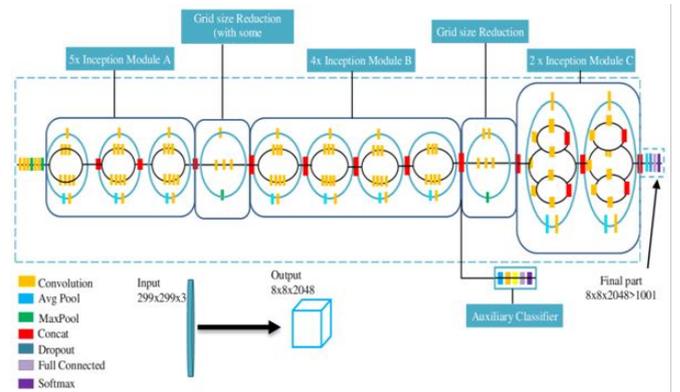

Fig. 7: Architecture of InceptionV3

MobileNetV2: MobileNetV2 is a mobile embedded vision convolutional neural network. MobileNetV2 reduces computational cost and maintains accuracy using depthwise separable convolutions after 3x3 convolutions. A depthwise separable convolution uses a single filter per input channel and a 1x1 pointwise convolution to construct new features from the outputs. This network-wide technique efficiently processes high-dimensional input. Pointwise convolutions extend the number of channels, depthwise convolutions, and pointwise convolutions project features back to a lower dimension in each inverted residual block with linear bottlenecks. A global average pooling layer reduces spatial dimensions to 1x1, a fully connected layer, and softmax classification output complete the network. Economical architecture with minimal latency and high precision is suitable for resource-constrained systems.

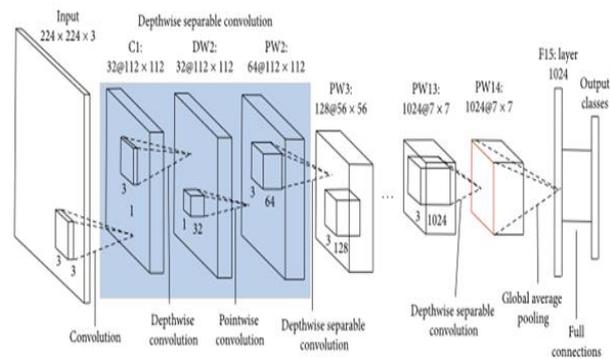

Fig. 8: Architecture of MobileNetV2

ResNet50: Image shows ResNet-50, a deep convolutional neural network that trains deep networks via residual learning. To reduce 224x224x3 input image to 112x112, a 7x7 convolutional layer (Conv1) with 64 filters is used. The architecture comprises four main phases (Layer1–Layer4) with several leftover blocks. These blocks simplify gradient backpropagation by bypassing convolutional layers with shortcut connections. Addressing the vanishing gradient problem permits deeper network training. Each layer deepens and shrinks the network. Layer1 has 64 filters, Layer2 128, Layer3 256, and Layer4 512, reducing spatial dimensions to 7x7. Features maps are flattened into 512-dimensional vectors and classified using a fully connected (Dense) and softmax layer after the final convolutional layer. With its depth and processing efficiency, ResNet-50 is a popular picture categorization system.

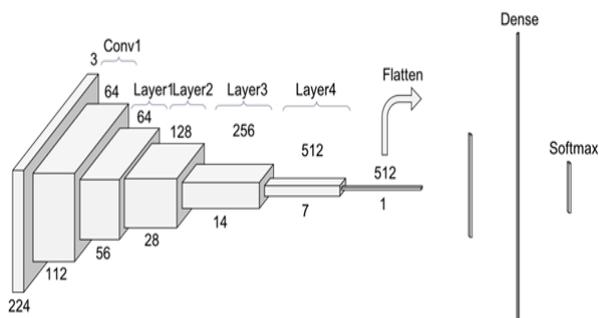

Fig. 9: Architecture of ResNet50

VGG16: VGG-16 architecture is a deep convolutional neural network noted for its simplicity and image categorization performance. VGG-16 uses modest 3x3 filters throughout its convolutional layers to process a 224x224x3 input image. The architecture has five blocks and 16 weight layers. Each block begins with one or more convolutional layers, increasing network depth. The first block has 64 filters, the second 128, the third 256, and the fourth and fifth 512. After each convolutional block, a max-pooling layer with a 2x2 filter and stride 2 halves spatial dimensions, minimizing overfitting and computational overhead. After the convolutional blocks, the network flattens the feature maps and passes them via three fully connected (FC) layers of 4096 units each, followed by a 1000-unit classification layer. The final softmax layer outputs the input image's 1000 class probabilities. This simple architecture has proven influential due to its successful usage of deep convolutional layers and narrow receptive fields, which perform well across image recognition tasks.

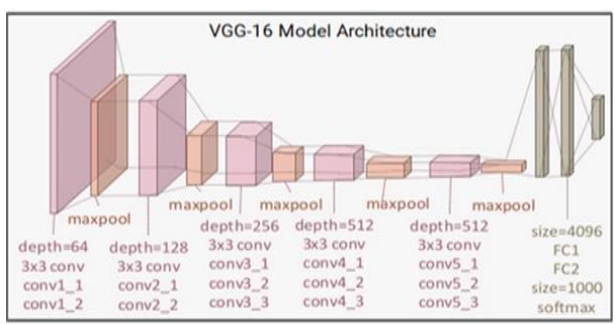

Fig. 10: Architecture of VGG19

VGG19: VGG-19 is a simple and deep convolutional neural network used in image categorization. To VGG-16, VGG-19 adds 19 weight layers—16 convolutional and 3 fully linked. The network uses five convolutional blocks for 224x224x3 input. Each block has many convolutional layers using 3x3 filters with 1 stride and 1 padding for spatial resolution. The first block has 64 filters, the second 128, the third 256, and the fourth and fifth 512. To reduce spatial dimensions, a 2x2 filter max-pooling layer with a stride of 2 follows each convolutional block. VGG-19 can learn more complex features than VGG-16 since each block has additional convolutional layers. After the convolutional layers, the network has three fully connected layers: two 4096-unit layers and one 1000-unit output class count layer. This architecture ends with a classification softmax layer. VGG-19's deep and simple design uses small 3x3 convolutional filters uniformly, making it successful for image recognition despite its higher computing cost.

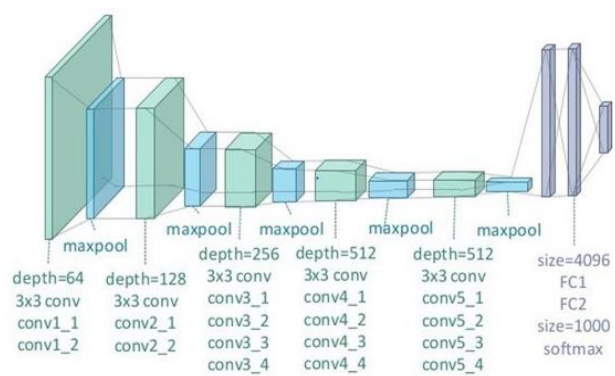

Fig. 11: Architecture of VGG19

3.4.3 Layers and Parameters

Each model has convolutional, pooling, fully connected, and batch normalization layers. The convolutional layers recognize patterns in the input images. They create feature maps by sliding learnable filters over the input image. Pooling layers reduce feature map spatial dimensions, reducing parameters and computational effort. The fully connected layers at the end of the network classify the features extracted by the convolutional layers. Most models use ReLU (Rectified Linear Unit), which introduces non-linearity. Certain models may use sigmoid or softmax activation functions in the output layer to generate classification probabilities. This method normalizes layer inputs to reduce internal covariate shift, stabilizing and speeding up training.

3.4.4 Hyperparameters

The performance of deep learning models is highly dependent on the choice of hyperparameters. The key hyperparameters used in training these models include:

Learning Rate: The learning rate controls how much the model's weights are adjusted with respect to the loss gradient. Typical values used range from 0.001 to 0.0001, depending on the model and dataset and in this experiment learning rate was taken as 0.001.

Batch Size: Batch size refers to the number of samples processed before the model's weights are updated. A batch size of 32 or 64 is commonly used, balancing computational efficiency with training stability and in the given experiment batch is equal to 32.

Epochs: The number of epochs refers to how many times the learning algorithm works through the entire training dataset. For this study, models were trained for 50-100 epochs, depending on the model's convergence. Due to the resource shortage 15 epochs were used in the experiment.

Optimizer: The choice of optimizer, such as Adam or SGD (Stochastic Gradient Descent), affects how the model's weights are updated during training. Adam is often preferred due to its adaptive learning rate capabilities and it was chosen for this experiment also.

3.4.5 Training Process

Training Procedure:

The models were trained using the Adam optimizer, which is widely used for its adaptive learning rate and efficiency in handling sparse gradients. After receiving batches of training data, the optimizer altered model weights to minimize loss function.

Loss Function:

The categorical cross-entropy loss function is ideal for multi-class classification. The projected probability distribution minus the actual distribution (true labels) is measured by this loss function. The goal during training is to minimize this loss to improve model accuracy.

3.5 Model Evaluation

Model performance must be assessed to determine its efficacy and make improvements. This study evaluated model performance using different metrics:

3.5.1 Metrics Used

Accuracy: Accuracy is the percentage of correctly classified cases. It is simple but may not accurately show model performance in imbalanced datasets. Common accuracy equation:

$$Accuracy = \frac{TP+TN}{TP+TN+FP+FN} \quad (1)$$

where,
$TP$ = True positive which refers to correctly predicted positive cases
$TN$ = True Negative which refers to correctly predicted negative cases
$FP$ = False positive which refers to incorrectly predicted positive cases
$FN$ = False negative which refers to incorrectly predicted negative cases

Precision: Precision measures the model's genuine positive predictions to its total positive predictions. It's useful when false positives are expensive. This formula calculates precision:

$$Precision = \frac{TP}{TP+FP} \quad (2)$$

Recall (Sensitivity): Recall measures the ratio of true positive predictions to the total actual positives. It is important when the cost of false negatives is high, such as in medical diagnosis tasks. Recall is calculated by the following formula:

$$Recall = \frac{TP}{TP+FN} \quad (3)$$

F1-Score: The F1-score is the harmonic mean of precision and recall, providing a single metric that balances both concerns. It is particularly useful in cases where the class distribution is uneven. F1-Score calculated by the following formula:

$$F1 - Score = 2 * \frac{Precision*Recall}{Precision+Recall} \quad (4)$$

## 4. RESULT ANALYSIS

This section illustrates the performances of the models those were applied in the experiments.

4.1 Comparison of Baseline Models

In order to illustrate the analysis of the performances of the models, their accuracy were taken into account first. Table 2 presents the training, validation and testing accuracy of the baseline models.

Table 2: Comparison of baseline models

|  | Training Accuracy | Validation Accuracy | Testing Accuracy |
|---|---|---|---|
| DenseNet121 | 64.83 | 72 | 69 |
| EfficientNetB0 | 57.2 | 66.98 | 66 |
| InceptionResNetV2 | 61.85 | 67.1 | 67 |
| InceptionV3 | 87.35 | 91.95 | 90 |
| MobileNetV2 | 97.89 | 98.13 | 98 |
| ResNet50 | 56.21 | 66.48 | 66 |
| VGG16 | 87.22 | 89.14 | 90 |

Based on the table provided, we can analyze the performance of various deep learning models used for keratoconus detection. Starting with DenseNet121, the model shows a noticeable improvement in accuracy from training (64.83%) to validation (72%), suggesting that the model benefits from regularization or other training strategies. However, the testing accuracy drops slightly to 69%, indicating potential issues with generalization. This drop could imply that the model might be slightly overfitting to the validation set or that it struggles to perform consistently on entirely unseen data.

EfficientNetB0 exhibits the lowest training accuracy at 57.2%, which could suggest underfitting during the training phase. Although its validation (66.98%) and testing (66%) accuracies are closer, its performance is lower than other models. This suggests that EfficientNetB0 needs more tuning or data to learn. InceptionResNetV2 improves somewhat from training (61.85%) to validation (67.1%) and maintains 67% testing accuracy. Like EfficientNetB0, its performance is disappointing compared to other models in the study.

Top performers InceptionV3 and VGG16 had training accuracies of 87.35% and 87.22%, respectively. Both models have high validation (91.95% for InceptionV3 and 89.14% for VGG16) and testing accuracies (90% for both) indicating good generalization. These models are reliable for classification because they generalize well to new data and have a narrow gap between validation and testing accuracies. These models balance complexity and performance, making them suitable for this study's dataset.

With 97.89%, 98.13%, and 98% training, validation, and testing accuracies, MobileNetV2 is the best model overall. MobileNetV2's accuracy consistency from training to testing implies it is neither overfitting nor underfitting and has good generalization. This model is the most reliable for real-world use due to its efficiency and precision.

However, ResNet50 and VGG19 perform badly, with ResNet50 having the lowest training accuracy (56.21%) and testing accuracy (66%). Despite being deeper than VGG16, VGG19 has lower training and testing accuracy of 61.94% and 68%, respectively. These results show that ResNet50 and VGG19 may need more optimization or may not be as well-suited to the dataset as MobileNetV2 and InceptionV3.

### 4.2 Error Analysis

The table compares deep learning models for keratoconus identification by precision, recall, and F1-score. Starting with DenseNet121, the model detects keratoconus with good precision (0.94), recall (0.92), and F1-score (0.93). Classifying normal and questionable instances drastically reduces its performance. The typical class has a moderate F1-score of 0.70 because to its poor precision (0.54) and high recall (0.98). DenseNet121 may be too sensitive, misclassifying other classes as normal. Suspect cases perform even worse, with an F1-score of 0.13, showing that the model struggles to correctly identify them, which may lead to frequent misclassifications.

EfficientNetB0 has similar issues. For keratoconus detection, it has high precision (0.98) and F1-score (0.93), but sharply lowers for normal and questionable cases. Another strong recall (0.99) but poor precision (0.51) gives the typical class a less effective F1-score of 0.67. High recall and low precision show EfficientNetB0 is prone to false positives, especially in normal class. With an F1-score of 0.04, the model rarely properly identifies questionable cases.

In comparison, InceptionV3 and MobileNetV2 excel. InceptionV3 scores well in all classes, especially the suspect class (F1-score of 0.83), which most models struggle with. Its high precision (0.97) and recall (0.98) for keratoconus identification and balanced performance in normal and questionable cases make it a reliable model. MobileNetV2 scores near-perfect across all parameters and classes, solidifying its top spot. As well as keratoconus detection (F1-score of 0.99), it is accurate in normal (0.98) and questionable cases (0.96). This consistency across all classes shows MobileNetV2's improved generalization and classification model robustness.

ResNet50 and VGG19 perform poorly in questionable cases. ResNet50's F1-score of 0 indicates that it cannot classify questionable cases. The lack of balance in handling all classes makes it less dependable, yet it detects keratoconus and normal cases well. VGG19 performs well in keratoconus detection (F1-score of 0.89), but it falters in the normal class (0.58) and fails to categorize questionable cases (0.14) like ResNet50.

MobileNetV2 is the best model for keratoconus identification since it balances precision, recall, and F1-score across all classes. InceptionV3 excels in the suspect class, which many models struggle with. However, ResNet50 and VGG19 have substantial shortcomings, notably in identifying questionable situations, emphasizing the need for accurate models that can manage imbalanced classes.

Table 3: Comparison of Precision, Recall and F1-Score

| Model | Class | Precision | Recall | F1-Score |
|---|---|---|---|---|
| DenseNet121 | Keratoconus | 0.94 | 0.92 | 0.93 |
|  | Normal | 0.54 | 0.98 | 0.7 |
|  | Suspect | 0.55 | 0.08 | 0.13 |
| EfficientNetB0 | Keratoconus | 0.98 | 0.88 | 0.93 |
|  | Normal | 0.51 | 0.99 | 0.67 |
|  | Suspect | 0.31 | 0.02 | 0.04 |
| InceptionResNetV2 | Keratoconus | 0.95 | 0.9 | 0.92 |
|  | Normal | 0.52 | 1 | 0.68 |
|  | Suspect | 0.5 | 0.02 | 0.04 |
| InceptionV3 | Keratoconus | 0.97 | 0.98 | 0.97 |
|  | Normal | 0.88 | 0.88 | 0.88 |
|  | Suspect | 0.83 | 0.82 | 0.83 |
| MobileNetV2 | Keratoconus | 0.99 | 0.99 | 0.99 |
|  | Normal | 0.98 | 0.97 | 0.98 |
|  | Suspect | 0.96 | 0.97 | 0.96 |
| ResNet50 | Keratoconus | 0.91 | 0.9 | 0.91 |
|  | Normal | 0.51 | 0.98 | 0.67 |
|  | Suspect | 0 | 0 | 0 |
| VGG16 | Keratoconus | 0.95 | 0.98 | 0.97 |
|  | Normal | 0.85 | 0.92 | 0.89 |
|  | Suspect | 0.88 | 0.75 | 0.81 |
| VGG19 | Keratoconus | 0.91 | 0.87 | 0.89 |
|  | Normal | 0.58 | 0.96 | 0.72 |
|  | Suspect | 0.41 | 0.08 | 0.14 |

## 4.3 Performance visualization

### 4.3.1 Learning Curves:

The figures above show deep learning model training and validation learning curves and their performance across epochs.

Training and Validation Loss

Training loss drops dramatically in the first few epochs for all models and subsequently plateaus, demonstrating that the models are learning from the data. Though models decrease validation loss at different rates, the trend is similar. MobileNetV2 and InceptionV3 have well-regularized validation losses that are lower than training losses, indicating that they do not overfit the training data. In contrast, EfficientNetB0 and ResNet50 exhibit a progressive decrease in validation loss, with some variation in later epochs, which may signal overfitting or generalization issues.

Training and Validation Accuracy

Training and validation accuracy curves reveal model performance. MobileNetV2 and InceptionV3 learn well, with training and validation accuracies rising to high values in the final epochs. This convergence shows these models can generalize to validation data without overfitting. These models have the best epoch markers near the conclusion of training, indicating consistent performance.

EfficientNetB0 and ResNet50 have a slower accuracy improvement and a large gap between training and validation accuracy, suggesting overfitting or underfitting. These models' validation accuracy curves do not improve as dramatically as their training accuracy, suggesting they are not properly capturing the validation data's patterns.

With constant loss and accuracy performance, MobileNetV2 and InceptionV3 are among the most effective models in this comparison. These models are suitable for real-world applications due to low overfitting and excellent validation accuracy. However, EfficientNetB0 and ResNet50 may need early stopping or data augmentation to promote generalization and minimize overfitting. Top models' learning curves are consistent, demonstrating their durability and ability to learn from data.

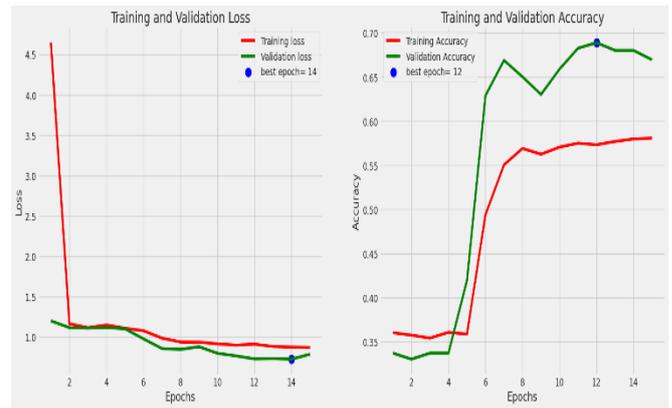

Fig 13. Training, validation Loss and Accuracy of EfficientNetB0

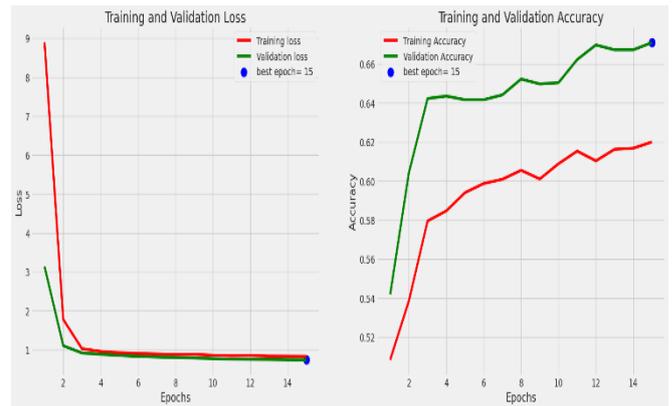

Fig 14. Training, validation Loss and Accuracy of InceptionResNetV2

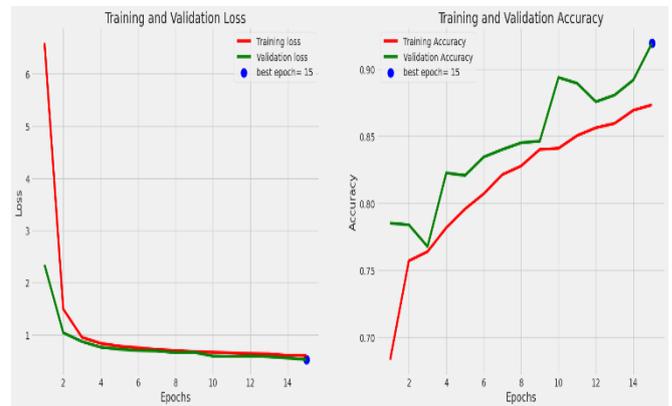

Fig 15. Training, validation Loss and Accuracy of InceptionV3

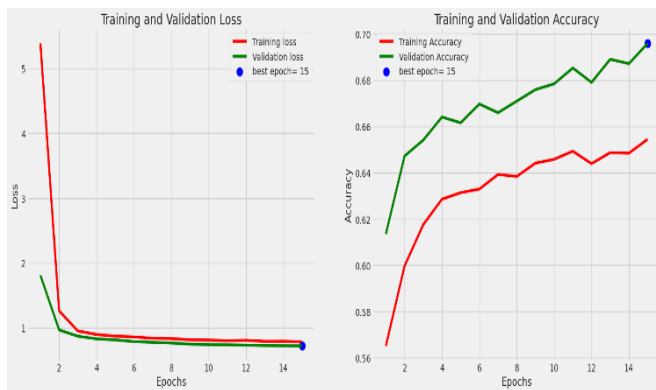

Fig 12. Training, validation Loss and Accuracy of DenseNet121

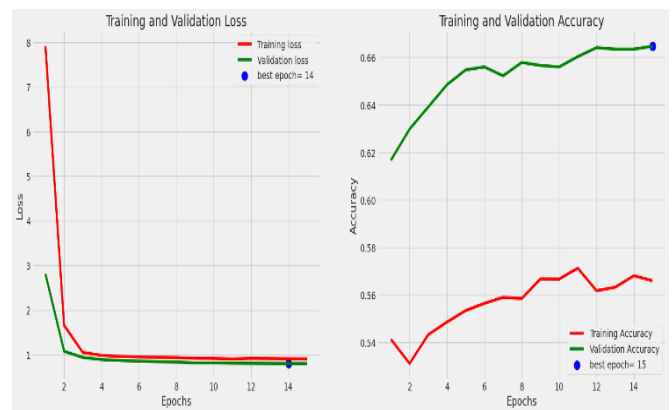

Fig 16. Training, validation Loss and Accuracy of MobileNetV2

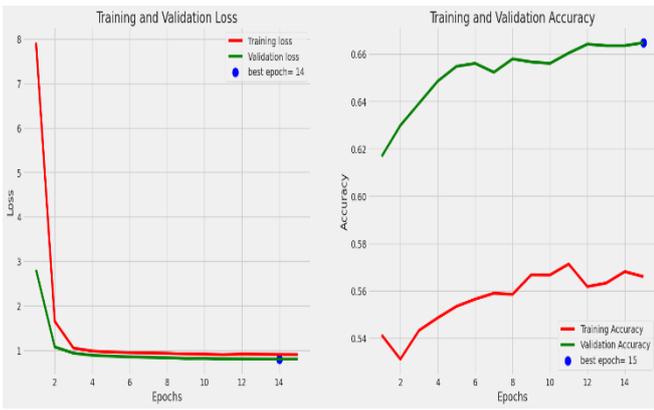

Fig 17. Training, validation Loss and Accuracy of ResNet50

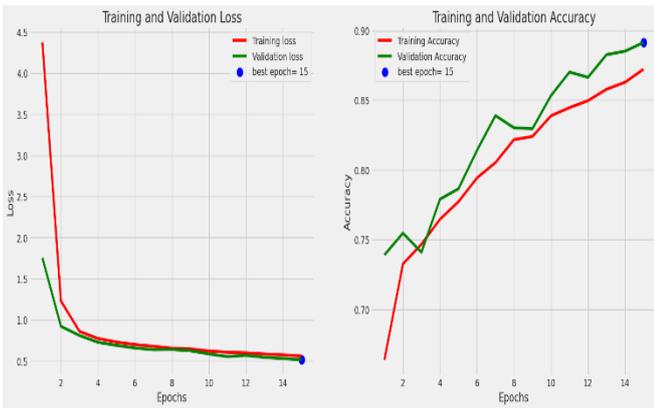

Fig 18. Training, validation Loss and Accuracy of VGG16

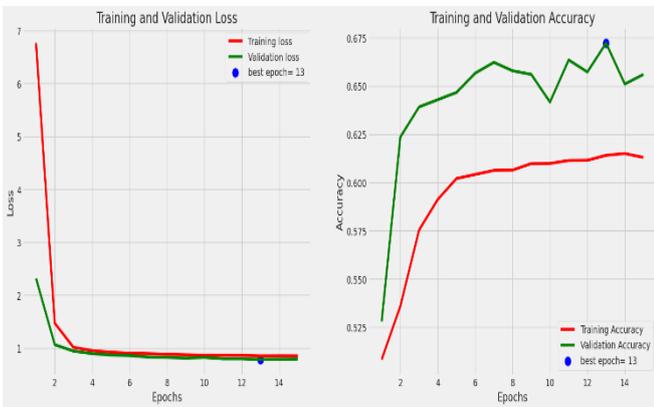

Fig 19. Training, validation Loss and Accuracy of VGG19

4.3.2 Confusion Matrices

The confusion matrices for different deep learning algorithms compare their keratoconus, normal, and suspect detection rates. Each matrix shows the models' correct classifications and errors, revealing their strengths and limitations.

Starting with DenseNet121, the model demonstrates a solid ability to identify keratoconus cases, correctly classifying 557 out of 603 instances. However, it shows some confusion between the keratoconus and suspect classes, as it misclassifies 23 keratoconus cases as normal and another 23 as suspect. The model is highly accurate in identifying normal cases, with 514 correct predictions, but there are still a few misclassifications, including 3 cases labeled as keratoconus and 7 as suspect. Notably, DenseNet121 struggles the most with suspect cases, where only 36 out of 475 are correctly identified, and a significant number (407) are incorrectly classified as normal, indicating a tendency to misidentify suspect cases as normal.

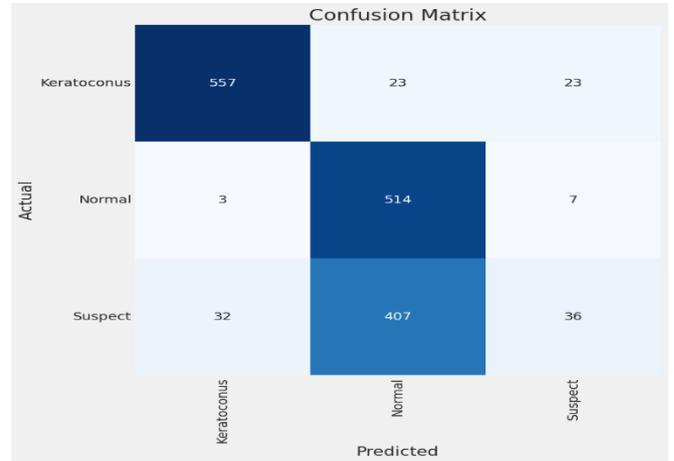

Fig 20. Confusion matrix of DenseNet121

EfficientNetB0 also detects normal cases well, identifying 521 of 524. Compared to DenseNet121, it misclassifies keratoconus and suspicious cases more often. The model misclassifies 52 keratoconus instances as normal and 22 as suspicious, demonstrating its difficulty separating these disorders. In the suspect class, EfficientNetB0 properly identifies only 11 out of 475 cases, misclassifying the rest as normal, highlighting its inability to generalize across case types.

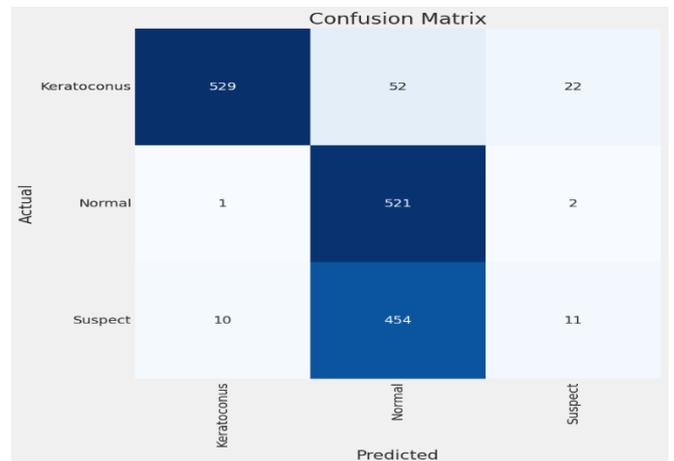

Fig 21: Confusion Matrix of EfficientNetB0

More balanced performance across classes is seen in InceptionResNetV2. The model correctly finds 541 keratoconus instances but misclassifies 52 as normal. Identifying 522 of 528 normal cases is highly accurate. It has trouble with questionable cases, mislabeling 438 of 475 as normal. This implies that InceptionResNetV2 can recognize keratoconus and normal cases, but it struggles to distinguish questionable cases from normal ones, resulting in frequent misclassifications.

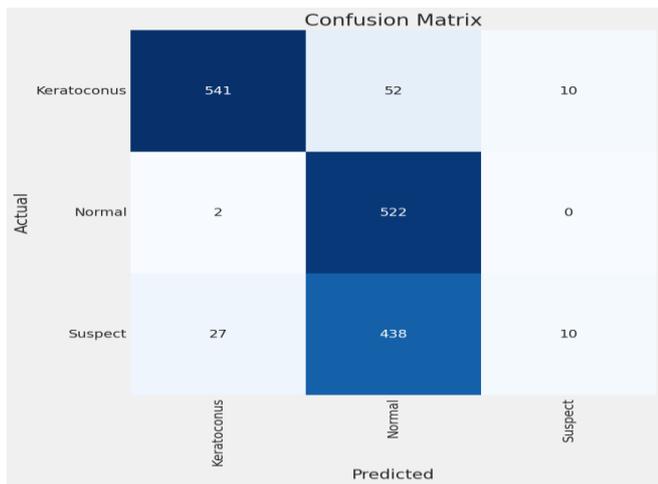

Fig. 22: Confusion matrix of InceptionResNetV2

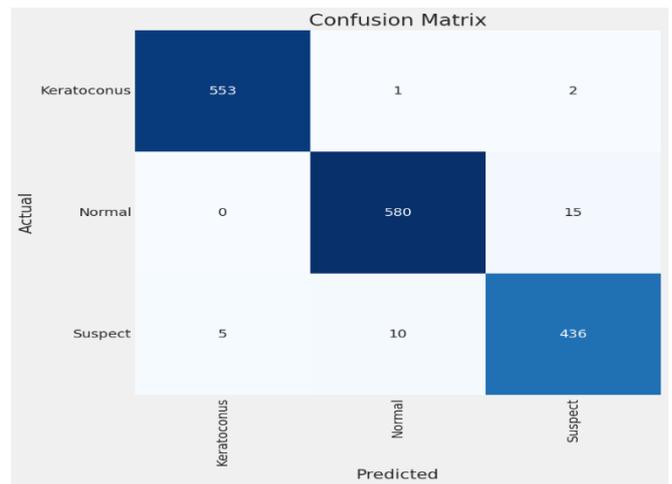

Fig. 24: Confusion matrix of MobileNetV2

However, InceptionV3 detects keratoconus and normal patients with little misclassifications. It accurately diagnoses 543 keratoconus instances, misclassifying 7 as normal and 6 as suspicious. The model predicts 525 of 528 normal situations correctly. However, it struggles with questionable cases, where 369 are correctly identified but 67 are misclassified as normal, indicating some uncertainty within this class, but less than in previous models.

Performance of the ResNet50 seems struggle to identify suspect from routine cases. The model accurately identifies 544 keratoconus instances, however only 432 of 475 suspicious cases are successfully recognized as normal. Additionally, 43 questionable cases are mistakenly categorized as keratoconus, showing that ResNet50 struggles with these more ambiguous cases and misclassifies more often than other models.

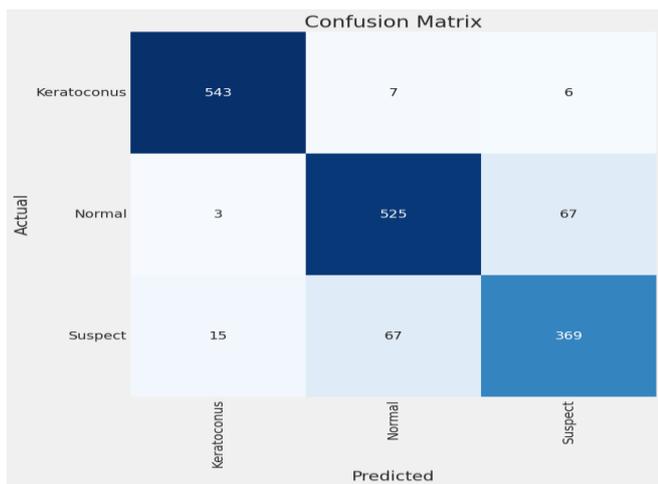

Fig. 23: Confusion matrix of InceptionV3

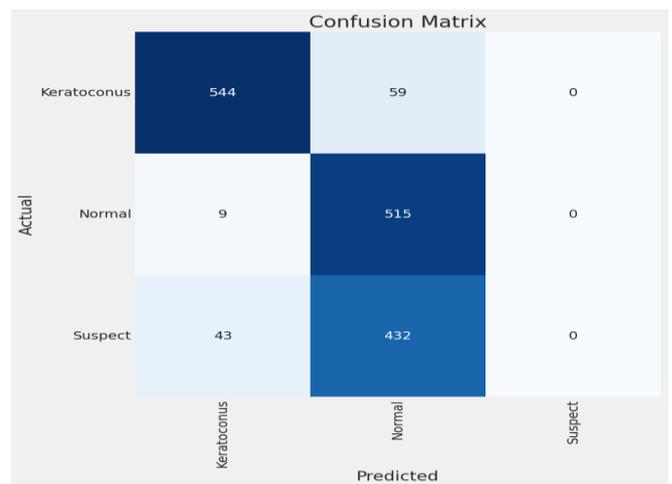

Fig. 25: Confusion matrix of ResNet50

MobileNetV2 stands out as the best-performing model overall, as reflected in its confusion matrix. It demonstrates near-perfect precision in identifying keratoconus cases, correctly classifying 553 out of 556, with only 1 misclassified as normal and 2 as suspect. The model also shows exceptional accuracy in identifying normal cases, with all 580 cases correctly classified, indicating no confusion with other classes. In the suspect class, MobileNetV2 still performs admirably, correctly identifying 436 out of 451 cases, though it misclassifies 10 as normal and 5 as keratoconus, making it the most balanced and effective model among those evaluated.

VGG16 and VGG19 exhibit similar challenges. VGG16 shows good accuracy in identifying keratoconus and normal cases but faces difficulties with the suspect class, where 339 out of 451 are correctly identified, but 86 are misclassified as normal. VGG19, while performing well in detecting normal cases (572 correctly identified out of 595), struggles with keratoconus and suspect cases. The model correctly classifies 483 keratoconus cases but misclassifies 39 as normal and 34 as suspect, showing that it has difficulty distinguishing between these conditions. Moreover, in the suspect class, VGG19 misclassifies 373 out of 451 as normal, highlighting significant challenges in its ability to accurately classify suspect cases.

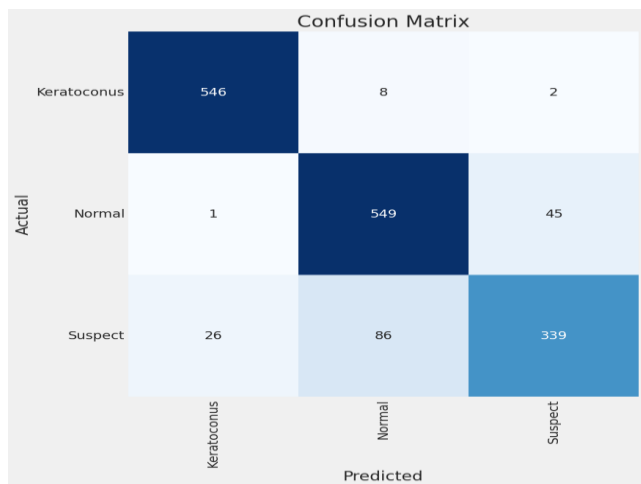

Fig. 26: Confusion matrix of VGG16

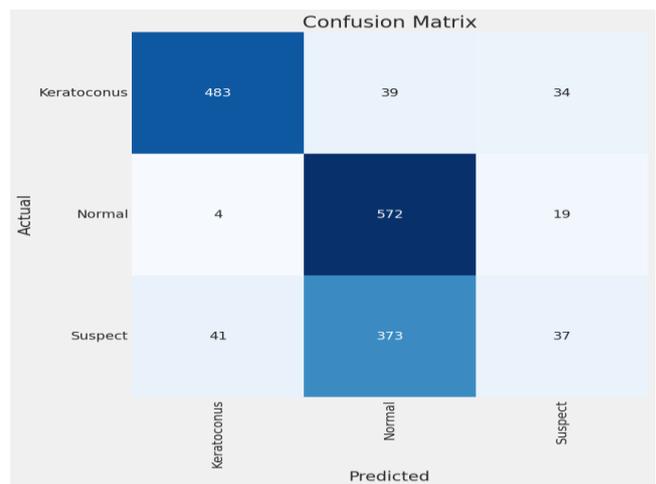

Fig. 27: Confusion matrix of VGG19

In summary, the confusion matrices emphasize that MobileNetV2 is the most accurate and balanced model across all classes, with minimal misclassifications. InceptionV3 and DenseNet121 also perform well, particularly in keratoconus and normal case detection, but they encounter more challenges with suspect cases. Models like EfficientNetB0, ResNet50, and VGG19 exhibit more significant difficulties, especially in accurately classifying suspect cases, which suggests areas for potential improvement through additional training data or model refinement.

## 5. CONCLUSIONS

A dataset of keratoconus, normal, and suspect cases is used to compare pre-trained Convolutional Neural Networks (CNNs) for keratoconus detection. The study shows that different models identify these circumstances with different accuracy and resilience. MobileNetV2 performed best, with excellent accuracy, precision, and recall across all classes and few misclassifications. InceptionV3 and DenseNet121 helped detect keratoconus and normal cases, but they struggled with questionable situations. However, models like EfficientNetB0, ResNet50, and VGG19 struggled to distinguish questionable cases from regular ones, revealing opportunities for improvement.

The findings suggest that enhanced CNN architectures could improve early diagnosis and treatment planning by automatically detecting keratoconus. The results also show that CNNs are powerful tools, but their performance depends on dataset complexity and small distinctions between classes, such as suspect and normal eyes. To achieve high diagnosis accuracy in clinical settings, model selection and training must be carefully considered.

## 6. FUTURE WORK

This research suggests numerous ways to improve deep learning model keratoconus detection accuracy and reliability. More advanced data augmentation techniques and more data sources are needed to help models distinguish between suspect and typical cases. To better understand corneal structures, the dataset could be supplemented by OCT or topographical maps from different devices.

In difficult instances, hybrid models that combine CNN strengths with other machine learning approaches like Recurrent Neural Networks (RNNs) or ensemble methods may increase classification accuracy. Integrating domain knowledge with clinical factors and imaging data may improve model performance. Finally, testing these models in clinical settings and longitudinally assessing patient outcomes would be helpful. Such attempts would test the models' practicality and reveal the long-term benefits of AI-driven keratoconus identification and control.